\pdfoutput=1
\documentclass[11pt]{article}

\usepackage{acl}

\usepackage{times}
\usepackage{latexsym}

\usepackage[T1]{fontenc}
\usepackage[utf8]{inputenc}

\usepackage{microtype}

\usepackage{algorithm}
\usepackage{algorithmic}
\usepackage{amsmath}
\usepackage{mathtools}
\usepackage{booktabs}
\usepackage{bm}
\usepackage{multirow}
\usepackage{hyphenat}
\usepackage{graphics}
\usepackage{caption}
\usepackage{tabularx}
\usepackage{svg}

\usepackage{verbatim} 
\usepackage{color}

\title{Grounding in social media: An approach to building a \\ chit-chat dialogue model }

\newcommand{\linecell}[2][c]{%
  \begin{tabular}[#1]{@{}c@{}}#2\end{tabular}}

\author{Ritvik Choudhary \\
  Waseda University \\
  \texttt{ritvik@fuji.waseda.jp} \\\And
  Daisuke Kawahara \\
  Waseda University \\
  \texttt{dkw@waseda.jp} \\}

\begin{document}
\maketitle

\begin{abstract}
Building open-domain dialogue systems capable of rich human-like conversational ability is one of the fundamental challenges in language generation. However, even with recent advancements in the field, existing open-domain generative models fail to capture and utilize external knowledge, leading to repetitive or generic responses to unseen utterances. Current work on knowledge-grounded dialogue generation primarily focuses on persona incorporation or searching a fact-based structured knowledge source such as Wikipedia. Our method takes a broader and simpler approach, which aims to improve the raw conversation ability of the system by mimicking the human response behavior through casual interactions found on social media. Utilizing a joint retriever-generator setup, the model queries a large set of filtered comment data from Reddit to act as additional context for the seq2seq generator. Automatic and human evaluations on open-domain dialogue datasets demonstrate the effectiveness of our approach.
% We measured significant improvements in engagement, relevance, and diversity of generated responses over a strong baseline.
\end{abstract}

\section{Introduction}
\label{intro}
Humans have long wanted to talk with the machine and have them comprehend and generate natural language. The task of chit-chat dialogue response generation can be described as one of the major goals in natural language processing. As such, there has been considerable interest in the sub-field of open-domain dialogue models.

Nevertheless, the existing dialogue response generation models still suffer from some very fundamental problems: lack of interesting (``Ok'', ``I see", etc.) or uninformative responses (``I don't know") (\citealp{li-etal-2016-diversity},~\citealp{shao-etal-2017-generating},~\citealp{ghazvininejad2017knowledge}). The primary cause for this is that, unlike humans, the models do not have access to knowledge, experience about out-of-domain topics or human conversational habits and hence can only produce limited unengaging generic responses.

Recent work has proposed considering additional context information such as multi-turn conversational history~\cite{zhang-etal-2018-personalizing},  persona~\cite{li-etal-2016-persona} or a fact-based knowledge base~\cite{dinan2018wizard}. Among these, our work approaches this problem from a more general standpoint of improving the raw conversational ability of generative models. We attempt this by taking inspiration from how humans learn to converse, i.e., through mimicking social interactions. Applying this in the context of dialogue models, we use a human-readable external knowledge base consisting solely of unstructured \textbf{s}ocial \textbf{m}edia \textbf{i}nteractions (hereinafter referred to as SMIkb), which tends to include a more diverse language structure and hence improve generated responses.

For our approach, we jointly train a generator-retriever model where the retriever searches through pre-indexed SMIkb and feeds the related information together with the input utterance to the generative seq2seq model, allowing for additional context at the time of generation.

In particular, we utilize the Dense Passage Retriever proposed by~\citet{karpukhin-etal-2020-dense} on top of BART~\cite{lewis-etal-2020-bart} as our generational model trained on a mix of open-domain dialogue datasets, together with a collection of Reddit submissions and comments as our main source of social interactions.
Experiments showed that our approach outperformed the existing vanilla seq2seq baseline (BART) across all of the automatic and human evaluation metrics. By making use of interactions grounded in social media, the generated responses were not only more engaging but were also shown to be much more relevant and natural, thus establishing the effectiveness of our approach.
\section{Related Work}
\label{related}

\begin{figure*}[t]
  \includegraphics[width=\textwidth]{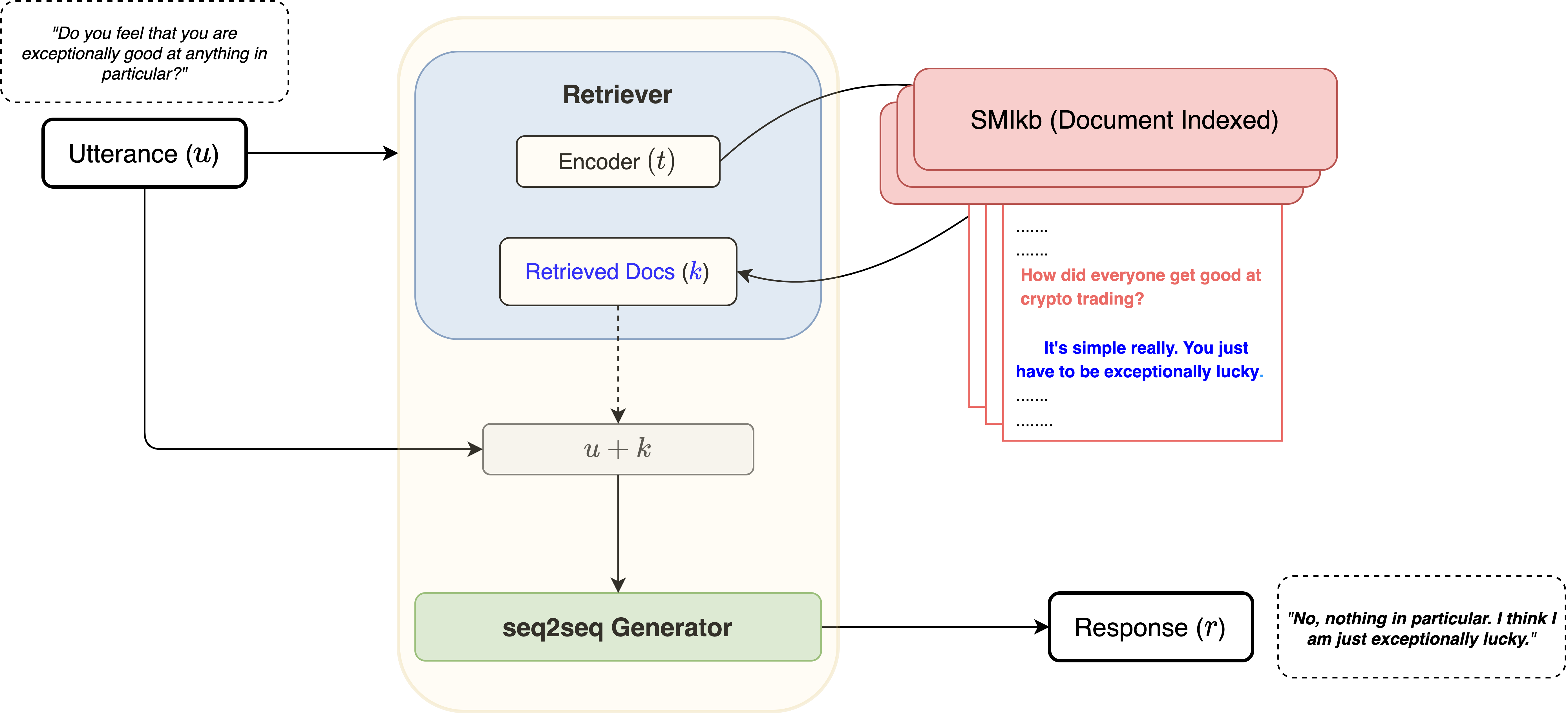}
  \caption{{Our proposed dialogue response generation approach grounded in SMIkb through a jointly trained retriever-seq2seq generator setup. Utterance $u$ is encoded and matched against titles (in $\textcolor{red}{\text{red}}$) where the respective comments ($k$, in $\textcolor{blue}{\text{blue}}$) are retrieved from the SMIkb. These act as an additional context for the generator to generate the final dialogue response $r$.}}
  \label{fig:model}
\end{figure*}

\paragraph{Dialogue Systems}
In recent years, major breakthroughs beginning with the Transformer~\cite{vaswani2017attention} and BERT~\cite{devlin-etal-2019-bert} have quickly shifted the landscape of modern NLP research. These were shortly followed by auto-regressive seq2seq models (T5 \cite{raffel2020exploring}, BART) that significantly improved performance on generation-based tasks such as dialogue systems. We adopt the widely accessible BART as our strong baseline.
% In the last decade, the approach of modeling conversations as a sequence to sequence (seq-2-seq) problem~\cite{sutskever2014sequence} became the backbone of modern generation-based dialogue systems with the advent of self-attention introduced in Transformer~\cite{vaswani2017attention}. Major breakthroughs came with BERT~\cite{devlin-etal-2019-bert} proposing a masked language modeling objective, and later auto-regressive seq2seq models (T5 \cite{raffel2020exploring}, BART, DialoGPT~\cite{zhang-etal-2020-dialogpt}) were proposed for generative tasks. We select the above-mentioned BART architecture to be our baseline.

\paragraph{Knowledge-based Conversational Models} Incorporating additional context or external information into existing models has been a field of much interest lately. Persona-chat~\cite{zhang-etal-2018-personalizing} or Empathetic Dialogues~\cite{rashkin-etal-2019-towards} take into account persona or empathetic information. Furthermore, advancements making use of knowledge bases in the area of open-domain dialogue systems have become increasingly common~\cite{ghazvininejad2017knowledge,dinan2018wizard}. The closest work to ours, in terms of including a retrieval step for dialogue generation, is~\citet{weston-etal-2018-retrieve}, which proposed an approach involving pre-training the retriever and generating only over the candidates retrieved in advance from the training set. More recently~\citet{roller-etal-2021-recipes} also tested retrieval-based dialogue generation. However, similar to~\citet{weston-etal-2018-retrieve}, they utilized a retrieval model that was kept fixed during training. Our work meanwhile follows a different direction that does not require pre-training of the retriever but fine-tunes it along with the generator to retrieve over a much larger knowledge base of interactions at generation time.

We would also like to mention~\citet{shuster2021retrieval}, which investigates factual hallucination in dialogue retrieval-generation models with a fact-based knowledge base such as Wikipedia. Our work takes a more generalized approach, focusing solely on improving the raw conversational ability of dialogue models. Instead of factual accuracy, we propose a simple approach for generating an engaging conversation grounded in unstructured social media interactions.
\section{Proposed Approach}
\label{app}
In this section, we discuss our approach to introducing social media interactions as an external knowledge base (SMIkb) to ground in for more natural and human-like response generation.
We begin with formulating the task of dialogue generation and then proceed to explain our joint retriever-generator model as the proposed setup for utilizing the aforementioned unstructured data source.
Note that in this work, we primarily focus on response generation for single-turn dialogues or dialogues. We decided that other settings such as a multi-turn case were best addressed in future work.

\subsection{Task Formulation}
\label{app:task}
Our task of response generation grounded in external knowledge can be formulated as training a model to predict a response $\mathbf{r}= (r_1,r_2,...,r_m)$ of $m$ words when given an input utterance $\mathbf{u}$ and a set of documents $\mathcal{D}$ that might contain relevant knowledge. We define our goal as to allow the model to learn the parameters such that when given an input utterance $\mathbf{u}$ and a knowledge base $\mathcal{D}$, the model can generate a response $\mathbf{r}$ following the probability $p(r_i | \mathbf{u}, \mathbf{r}_{<i}, \mathcal{D}; \theta)$, where $\theta$ refers to the parameters of the model.

\subsection{Model}
\label{app:model}

Inspired by recent advances in retrieval assisted QA~\cite{guu2020realm, lewis2021retrievalaugmented}, we adopt a simple joint retriever-generator setup to the task of dialogue generation.
Concretely, we utilize BART, a seq2seq model pre-trained on a denoising objective, as our generative model along with the pre-trained neural Dense Passage Retriever (DPR)~\cite{karpukhin-etal-2020-dense} as the retriever of choice. DPR is a highly efficient neural retriever pre-trained for retrieving the top-$\bm{k}$ similar documents to an input query $\bm{u}$. It executes this by encoding both the query and the entire knowledge base through independent BERT-based encoders (as $\bm{t}$). Furthermore, we follow~\citet{karpukhin-etal-2020-dense} to build an offline searchable dense vector index of these embeddings for our SMIkb using the FAISS~\cite{JDH17} library for faster lookup.
An overview of our architecture is shown in Figure \ref{fig:model}. Application of our model to dialogue response generation can be formulated as a two-step process: (1) the retriever searching top-$k$ documents from the pre-indexed interaction knowledge base, relevant to the input utterance, and (2) the generator predicting the response to the previous utterance along with the retrieved context.

Following the notion set in Section \ref{app:task}, the probability of generating the response $\mathbf{r}$ given the utterance $\mathbf{u}$ and each of the top-$k$ documents ${d}_j$ from the knowledge base $\mathcal{D}$ can be defined as
\begin{equation}
    \resizebox{\hsize}{!}{$p(\mathbf{r}|\mathbf{u};\theta,\lambda) = \sum_{j}^{k} p_{\lambda} (d_j|\mathbf{u};\lambda) \prod_{i} p_{\theta}(r_i | \mathbf{u}, \mathbf{r}_{<i}, d_j; \theta)$},
\end{equation}
where $\theta$ and $\lambda$ are parameters for the generator and retriever, respectively. They are both fine-tuned jointly in an end-to-end fashion, with the retriever providing additional context that is concatenated together with the input at the time of generation. As there is no ``correct'' document source in the knowledge base, we consider it to be a latent variable. Therefore, during decoding we marginalize these probabilities over all the retrieved documents to return the most probable (best) response using beam search.

\begin{table}[t]
	\centering
	\resizebox{\linewidth}{!}{%
	\begin{tabular}{lcrrr}
		\hline
	    Dataset & Total (turns) & Train & Valid & Test \\ \hline
		DailyDialog & 76,743  & 53,721 & 11,511 & 11,511 \\
		DailyDialog++ & 39,913  & 27,939  & 5,987 & 5,987 \\
		Cornell Movie-Dialogs & 221,088 & 154,762 & 33,163 & 33,163 \\ \hline
		Reddit (pseudo extracted) & 200,000 & 140,000 & 30,000 & 30,000 \\ \hline
	\end{tabular}%
	}
	%\vspace*{-1ex}
	\caption{{Overview of datasets in use.}}
	\label{tab:dialdata}
	%\vspace*{-1ex}
\end{table}
\section{Experiments}
\label{exp}
We evaluate our model together with various external knowledge datasets on a mixture of open-domain dialogue datasets. The results are then compared with two BART-based baselines.

\begin{table*}[t]
    \center
    \resizebox{\linewidth}{!}{%
    \small{
    \begin{tabular}{lllrrrrrrrrr}
    \toprule 
     Model Setup & Training &  Knowledge Base &  &  &  & {\small{BLEU-4}} & {\small{Dist-1}} & {\small{Dist-2}}\\
       & Data & (Retrieval)  \\ 
    \midrule 
    Baseline 1 & ODD & None &  &  &  & 1.31 & 0.20 & 0.96 &  &  &\\
    Baseline 2 & ODD + SMIkb & None & & & & 1.05 & 0.12 & 0.47 &  &  & \\
    \midrule
    & & & \multicolumn{3}{c}{$k=3$} 
      &  \multicolumn{3}{c}{$k=5$} & \multicolumn{3}{c}{$k=7$} \\
      & & & {\small{BLEU-4}}  & {\small{Dist-1}} & {\small{Dist-2}} & {\small{BLEU-4}} & {\small{Dist-1}} & {\small{Dist-2}} & {\small{BLEU-4}}  & {\small{Dist-1}} & {\small{Dist-2}}\\
    \midrule
    \textit{Ours} (SMIkb) & ODD & SMIkb & \textbf{9.78} & \textbf{2.80} & \textbf{16.90} & \textbf{\underline{10.51}} & \textbf{5.50} & \textbf{\underline{26.63}} & \textbf{10.48} & \textbf{\underline{5.51}} & \textbf{26.62}\\
    \textit{Ours} (Wiki) & ODD &  Wiki & 6.93 & 2.57 & 14.91  & 7.14 & 4.94 & 23.38 & 7.11 & 5.02 & 23.79\\
    \textit{Ours} (Mix) & ODD &  SMIkb + Wiki & 6.03 & 2.45 & 14.08  & {6.20} & 4.71 & {22.25} & 6.21 & 4.71 & {22.23}\\
    \bottomrule
    \end{tabular}}%
    }
    % \vspace*{-1ex}
    \caption{{Automatic evaluation of generated responses across various values of $k$ for top-$k$ document retrieval. The baselines do not have a retrieval step and therefore do not have an effect due to changing $k$.} {\textbf{\underline{bold}} refers to the best scores across all $k$ among the generated responses. ODD is the collection of \textbf{O}pen-\textbf{D}omain \textbf{D}atasets from Section 4.2.}}
    \label{tbl:res:both}
    % \vspace*{-1ex}
\end{table*}

\begin{table}[t]
    \center
    \resizebox{\linewidth}{!}{%
    \footnotesize{
    \begin{tabular}{lrrr}
    \toprule 
     Model Setup & \multicolumn{3}{c}{Human Eval.}  \\
      \cmidrule(rr){2-4}
        & {\small{Relevance}}  & {\small{Engagement}} & {\small{Knowledge}} \\ 
    \midrule 
    Gold (Test-Data) &  3.50 & {3.33} & {3.47} \\
    \midrule
    Baseline 1  & 2.82 & 2.35 & 3.00 \\
    Baseline 2  & 3.03 & 3.02 & 2.89 \\
    \midrule
    \textit{Ours} (SMIkb) & \textbf{3.84} & {3.75} & {3.60}\\
    \textit{Ours} (Wiki) &  3.40 & 3.75 & \textbf{3.76} \\
    \textit{Ours} (Mix) & 3.62 & \textbf{3.80} & 3.71 \\
    \bottomrule
    \end{tabular}}%
    }
    % \vspace*{-1ex}
    \caption{{Human evaluation of responses for the best $k=5$.}}
    \label{tbl:res:human}
    % \vspace*{-1ex}
\end{table}

\begin{table*}[t]
\centering
\resizebox{\linewidth}{!}{%
\small
\begin{tabular}{p{.185\linewidth} p{.185\linewidth}p{.185\linewidth}p{.185\linewidth}p{.185\linewidth}} \hline
        Input Utterance & SMIkb & Wiki & Mix & Baseline \\ \hline \\
        
        Here comes the birthday cake. & Wow, what a lovely cake. I'll have a piece. & Wow, what a beautiful cake. It must have tasted as good as I expected. & Really? Wow. Happy birthday to you! You look so healthy. &  Thank you. \\\\ \hline \\
        I'd better go. I gotta get up early. Go over to my parents'. & What are you going to do for breakfast? & You mean you don't wanna go with me? & Why don't you stay at my place? I'll cook dinner. & I don't think so. \\\\ \hline \\
        Something's burning? & Oh my goodness, it's on fire! & Oh my goodness, it ’ s burning! & Oh, my bread! It's burning! & No, it's not. \\\\ \hline \\
        Would you mind if we just called it an early night? & Even I couldn't. I have work tomorrow. & I'd love to.  But I've got to work tomorrow. & What are you talking about? We just ate? & I don't think so. \\\\ \hline \\
        Isn't it wonderful walking here? & What do you mean? Was it bad? & Not at all. The air is fresh, the walk is easy and the sunshine is kind. & Yes, you can see the whole city from here. & Yes, it's very beautiful. \\\\ \hline
    \end{tabular}%
    }
    \caption{A comparison of responses generated from various sources.}
    \label{tbl:gen-resp}
\end{table*}

\begin{table*}[t]
	\centering
	\resizebox{\linewidth}{!}{%
	\begin{tabular}{cc}
        \hline
		\multicolumn{2}{c}{SMIkb} \\ \hline
		\multicolumn{1}{c}{\linecell{title}} & {\linecell{text}} \\ \hline
		\multicolumn{1}{c||}{\linecell{LPT: If you borrow something like a tool or a generator\\from someone, return it in BETTER shape than you got it.}}   & \linecell{My dad always said that returning something in the\\same condition you received it is the absolute bare minimum.} \\ \hline
		\multicolumn{1}{c||}{\linecell{SoftBank Nears \$40 Billion Deal to Sell Arm Holdings to Nvidia}}   & \linecell{Nvidia is priced decentlyfor what they offer.} \\ \hline
		\multicolumn{1}{c||}{\linecell{Apple to Give Employees Paid Time Off to Vote in U.S. Election}}     & \linecell{This exactly. A large majority of disenfranchised\\communities work jobs that don't observe federal holidays.} \\ \hline
		\multicolumn{1}{c||}{\linecell{Apple may be working on a foldable iPhone}}     & \linecell{I can confirm that Apple would be stupid to not be working on one.\\Whether they ever release one is up for debate,\\but they're definitely working on one.} \\ \hline
	\end{tabular}%
	}
	\caption{Snapshot of SMIkb.}
	\label{tab:knowdata}
\end{table*}

\begin{table*}[t]
\centering
\resizebox{\linewidth}{!}{%
\small
\begin{tabular}{p{.185\linewidth} p{.185\linewidth}p{.185\linewidth}p{.185\linewidth}p{.185\linewidth}} \hline
        Input Utterance & SMIkb & Wiki & Mix & Baseline \\ \hline \\
        
        Who was the conductor? & Oh speaking of the conductor, he was indeed a genius. A descendent of Chinese. & I don't remember the name. & Oh, speaking of the conductor, he did an excellent job. &  I know. \\\\ \hline \\
        Oh god you scared me! & Ahahah! We need a fitness instructor. & I'm sorry. & Sorry. I was trying to scare you. & Is something wrong? \\\\ \hline \\
    \end{tabular}%
    }
    \caption{Selected examples where the proposed model struggles or fails at generating a relevant response.}
    \label{tbl:gen-resp-negative}
\end{table*}

\subsection{SMIkb}
\label{exp:kb}
Aiming to improve the raw communication ability of dialogue systems by mimicking human response behavior, we built our external knowledge base of unstructured social media interactions (SMIkb). It comprises of entries from top thread titles and their top 100 comments from Reddit, an American social news aggregation and discussion site, throughout 2020 (January-November). A total of 1.6 million entries were first scraped through the open-sourced Pushshift API~\cite{DBLP:conf/icwsm/BaumgartnerZKSB20} of which a random selection of 600,000 (due to memory limitations) makes up our SMIkb. A snapshot of the same is shared in Table \ref{tab:knowdata}.

Furthermore, to verify the effectiveness of using a conversational knowledge base like Reddit, we compared ours to a pure Wikipedia knowledge base (ref. ``Wiki'') of the same size (random sample of 600k entries) containing the wiki page title and the leading 100 words. Additionally, we also tested a 1:1 combination of the above two bases (ref. ``Mix'').

\subsection{Datasets}
\label{exp:data}
We fine-tune our models on a variety of open-domain and scraped dialogue datasets.

\vspace*{-0.5ex}

\paragraph{Open-domain datasets} We use a combination of DailyDialog \cite{li-etal-2017-dailydialog} and DailyDialog++~\cite{Sai_2020} as high-quality daily life-based dialogue sets. We also consider the Cornell Movie-Dialogs Corpus~\cite{danescu-niculescu-mizil-lee-2011-chameleons}, which is a corpus of scripts of movie dialogues.

\vspace*{-0.5ex}

\paragraph{Reddit} Furthermore we extract another 200,000 comment pairs from Reddit, distinct from the SMIkb, to act as a pseudo dialogue dataset to supplement our knowledge base.

An overview of the datasets is listed in Table \ref{tab:dialdata}.

% \begin{table*}[t]
%     \center
%     \resizebox{\linewidth}{!}{%
%     \footnotesize{
%     \begin{tabular}{lrrr}
%     \toprule 
%      Model Setup & {$k=3$} 
%       &  {$k=5$} &  {$k=7$} \\ 
%     % \midrule 
%     % Gold (Test-Data) &  - & - & - & 3.50 & \textbf{3.33} & \textbf{3.47}\\
%     \midrule
%     \textit{Ours} (SMIdb) &  {9.78} & \textbf{10.51} & {10.48}\\
%     \textit{Ours} (Wiki)  & 6.93 & 7.14 & 7.11\\
%     \textit{Ours} (Mix)  & 6.03 & {6.20} & 6.21\\
%     \bottomrule
%     \end{tabular}}%
%     }
%     \caption{{Automatic and human evaluation of generated responses.} \small{* refers to best scores among the generated responses (excluding gold). ODD is the collection of \textbf{O}pen-\textbf{D}omain \textbf{D}atasets from Section 4.2.}}
%     \label{tbl:res:both}
%     % \vspace*{-1ex}
% \end{table*}

% \begin{table*}[t]
%     \centering
%     \resizebox{\linewidth}{!}{%
%     \tabcolsep 3pt
%     \label{tab:res_ex}
%     \begin{tabular}{p{.3\linewidth}|p{.34\linewidth}p{.26\linewidth}p{.3\linewidth}}
%         \hline
%         Utterance & SMIdb & Wiki & Baseline \\
%         \hline
%         That Japanese restaurant, Shogun, got good reviews. & Oh yeah! I saw a review of that restaurant on television. & Really? Why did you want to go there? & I don't have any idea about that. \\\hline
%         An omelet? & I'm trying to be healthy, remember? & I'm not having an omelet. & No, no. \\
%         \hline
%     \end{tabular}%
%     }
%     \caption{{A sample of generated responses.}}
%     \label{res:tbl:resp}
% \end{table*}

\subsection{Experimental Setup}
\label{exp:setup}
\paragraph{Implementation Details} Our joint retriever-generator model consists of a pre-trained Dense Passage Retriever and BART-large (24 layers, 406M), which are later fine-tuned together on SMIkb and dialogue datasets. The model is trained mostly with the default parameters, batch size of 1, and an initial learning rate of $3\times10^{-5}$. We further experiment with various values of $k$ for our top-$k$ document retrieval, while beam search with size of $5$ is used as our response decoding strategy.
% Fine-tuning is performed with an Nvidia V100 GPU.

\paragraph{Baseline} We consider two strong baselines based on a vanilla BART-large with no retriever to investigate the effectiveness of our approach. The first is fine-tuned solely on the datasets mentioned in Section \ref{exp:data} (ref. ``Baseline 1'') with no SMIkb. Next to confirm the effectiveness of our providing external data through our retriever-generator setup, we merge the entire SMIkb interactions into our training data, and simply fine-tune the vanilla model on this new extended set. (ref. ``Baseline 2''). Note that although we choose BART as our generator and baseline for its size and relative ease in training, our proposed SMIkb based modeling setup could possibly also be extended to larger models.

\subsection{Evaluation}
To measure the impact of social media interactions, the generated responses were evaluated through both automatic and human evaluations. The results are compiled in Tables \ref{tbl:res:both} and \ref{tbl:res:human}.

% \begin{table}[t]
% \centering
% \resizebox{\linewidth}{!}{%
% \begin{tabular}{cccc}
% \hline
%     Model DB & BLEU-4 &Distinct-1 & Distinct-2 \\\hline
%     None (Baseline) & 1.31 & 0.20 & 0.96 \\
%     None (Baseline SMIdb merged) & 1.05 & 0.12 & 0.47 \\
%     \hline
%     Mix (SMIdb + Wiki) & 6.03 & 2.45 & 14.08 \\
%     Wiki & 6.93 & 2.57 & 14.91 \\
%     SMIdb (ours) & \textbf{9.78} & \textbf{2.80} & \textbf{16.90} \\\hline
% \end{tabular}%
% }
% \caption{Automatic evaluation of generated responses.}
% \label{tbl:res:auto}
% \end{table}

\paragraph{Automatic} We perform a series of automatic evaluations on the generated responses. Starting with the BLEU~\cite{papineni-etal-2002-bleu} score for relevancy, we also
calculate Distinct-N~\cite{li-etal-2016-diversity} for the amount of diversity introduced.

\paragraph{Human} It has been widely reported that automatic metrics do not sufficiently evaluate the actual quality of the generated outputs~\cite{liu-etal-2016-evaluate}. Thus, we additionally performed human evaluation of the responses with the highest BLEU ($k=5$) through Amazon Mechanical Turk, on the following three metrics: \textit{Relevance}, whether the response is relevant to the utterance; \textit{Engagement}, whether the response makes the conversation engaging; and \textit{Knowledge}, whether the response seems knowledgeable or sensible. 
% For carrying out the human-evaluation of the generated responses, we selected our evaluators to be English speakers from the United States with an approval rate of over 90\%. The average pay was set at $1$ cent per question.
The evaluators were asked to score 100 responses selected at random from the test set, on a scale of 1-5. Each response was scored by 7 different evaluators, and their average was calculated.
We selected our evaluators to be English speakers from the United States with an approval rate of over 90\%.

% Furthermore, we also requested the raters to select a preferred response in a direct comparison of our approach against the baselines. The results along with the rater-agreement compiled in Table.

% We did not consider reporting inter-annotator agreement as we believe it is not suitable for crowdsourcing-based evaluation, and in related works it is more common to take the average of crowdworkers' scores instead.

\subsection{Discussion}
\label{result}

First, with automatic evaluation, we observe that our method of introducing social interactions through a retriever at generation time maintains task performance and allows for a more diverse set of responses, as shown with an increase in all of the measured metrics over both the baselines. Moreover, our Reddit-based SMIkb model outperformed other combinations, confirming an increase in response quality. Furthermore, experiments comparing the effect of top-$k$ on generation quality showed $k=5$ as the sweet spot with the highest BLEU, which was then evaluated further.

Similar to the automatic evaluation, all of the proposed combinations showed improvements over the baseline in human evaluation.
In line with our initial hypothesis, the SMIkb model recorded the highest score in terms of conversational relevance. Meanwhile, the mix of SMIkb and Wikipedia seems to find the appropriate balance between conversational ability and ``knowledgeability'' with the metrics even exceeding the gold score. We hypothesize that this is not just due to high generation quality but also the nature of our task, which might make our model responses more relevant to a worker when assessed in a single-turn context.
Overall, the results verified our proposed approach of learning directly from social media interactions leading to large improvements over the baseline in all metrics of interest. Furthermore, significance tests of bootstrap re-sampling showed that the above evaluation score differences between baselines and our models were statistically significant ($p<0.01$). Examples of the various generated responses across all the models are shared in Table \ref{tbl:gen-resp}.

In addition, we looked at cases where our model struggles or fails at generating a natural response, a select few of which are compiled in Table \ref{tbl:gen-resp-negative}. In some of these particular cases, the baseline generations, although generic, turn out to be more relevant. We believe some of these long responses with unrelated information to be an artifact of our generation model being overly dependent on the knowledge base. 
While social media may simulate human-like conversations in a large variety of situations, it is still far from being a perfect stand-in for real-life dialogue. Therefore, our future work in this direction should look at not only the quality and scope of the knowledge base, but also consider selecting \textit{when} to ground and make use of the said knowledge during response generation.
\section{Conclusion}
\label{conc}
We aimed to improve the raw conversational ability of dialogue systems by grounding the responses in much more human-like social media interactions. Our approach involved a neural retriever-seq2seq generator model fine-tuned jointly, where relevant knowledge was retrieved at the time of generation to assist a more engaging and natural dialogue response. Our experiments 
showed significant improvements with both automatic and human evaluation metrics ranking the SMIkb-grounded replies to be much more diverse, engaging, and relevant.
%, proving the overall effectiveness of our simple method for open-domain dialogue response generation.

\section*{Acknowledgements}
This work was supported by a joint research grant from LINE Corporation.

\bibliography{anthology,custom}
\bibliographystyle{acl_natbib}

% \appendix

% \section{Example Appendix}
% \label{sec:appendix}

% This is an appendix.

\end{document}